\documentclass[10pt,twocolumn,letterpaper]{article}
\usepackage{preprint}
\newcommand{\scoreup}{\rlap{\hspace{1pt}\textcolor{blue!70!black}{\normalsize$\blacktriangle$}}}

\definecolor{linkblue}{rgb}{0.13,0.27,0.43}
\usepackage[breaklinks,colorlinks,allcolors=linkblue]{hyperref}
\usepackage[capitalize]{cleveref}
\hypersetup{
 pdftitle={ScoutCache: Efficient Few-Step Autoregressive Video Diffusion via Conditional Readout and Selective Caching},
 pdfauthor={Hao Tan, Jinghui Qin, Xuming Hu},
 pdfsubject={Few-step autoregressive video diffusion acceleration}
}
\title{ScoutCache: Efficient Few-Step Autoregressive Video Diffusion via Conditional Readout and Selective Caching}
\newcommand{\HaoTanEmail}{tanhao4869@gmail.com}
\newcommand{\JinghuiQinEmail}{scape1989@gmail.com}
\newcommand{\XumingHuEmail}{xuminghu@hkust-gz.edu.cn}

\newcommand{\ScoutCacheAuthorBlock}{%
\begin{minipage}{\textwidth}
\centering
{\fontsize{12}{14}\selectfont\bfseries
Hao Tan\textsuperscript{1},\quad
Jinghui Qin\textsuperscript{2},\quad
Xuming Hu\textsuperscript{1,3,*}\par}
\vspace{1pt}
{\fontsize{11}{11.5}\selectfont
\textsuperscript{1}The Hong Kong University of Science and Technology (Guangzhou)\\
\textsuperscript{2}Guangdong University of Technology\\
\textsuperscript{3}The Hong Kong University of Science and Technology\par}
\vspace{1pt}
{\fontsize{11}{11.5}\selectfont\rmfamily
\HaoTanEmail,\quad\JinghuiQinEmail,\quad\XumingHuEmail\par}
\vspace{1pt}
{\fontsize{11}{11.5}\selectfont
\textsuperscript{*}Corresponding author.\par}
\end{minipage}%
}

\author{\ScoutCacheAuthorBlock}
\date{}
\begin{document}
\maketitle
\begin{abstract}
Few-step autoregressive video diffusion enables efficient long video generation, but inference still incurs substantial computational costs from repeated denoising and KV cache operations. Existing acceleration methods reduce parts of this overhead, while historical KV reading and cache construction remain costly across video chunks. To address these challenges, we propose ScoutCache, a training-free framework that jointly optimizes denoising, historical KV reading, and cache construction. We observe that shallow feature changes provide useful information for estimating clean latent updates, while historical KV dependence varies across Transformer layers and attention heads but remains relatively stable across video chunks. Based on these observations, we develop Conditional Denoising Readout (CDR) to estimate intermediate denoising updates from shallow feature changes and known noise variations. We further introduce Output Risk Guided Historical KV Reading (ORHR) to allocate historical reading budgets based on attention output perturbations, and Content and Coverage Guided Sparse Cache Construction (CCSC) to reduce cache construction costs by selecting informative spatial tokens while maintaining spatial coverage. Extensive experiments on Self-Forcing and LongLive using VBench and MovieGen demonstrate that ScoutCache achieves up to 2.28× and 1.86× speedups, respectively, while consistently improving generation quality without modifying pretrained model parameters.
\end{abstract}
\section{Introduction}
\label{sec:intro}


Autoregressive video diffusion models have emerged as a promising approach to long-form video generation, synthesizing videos chunk by chunk while conditioning on previously generated content~\cite{huang2026self, yang2025longlive, chen2024diffusion, chen2025skyreels}. Recent advances in few-step diffusion have further improved generation efficiency by substantially reducing the number of denoising iterations~\cite{yin2025slow, liu2026rolling}. For example, Self Forcing~\cite{huang2026self} enables efficient autoregressive generation through few-step denoising and KV caching, while LongLive~\cite{yang2025longlive} supports real-time long-video generation with short-window attention and a frame-level attention sink. However, despite these advances, autoregressive inference still involves substantial computational overhead, particularly when maintaining historical context across video chunks~\cite{teng2025magi, gao2024ca2}. In the few-step KV-cached setting, each video chunk requires repeated Transformer evaluations for denoising, historical KV reads to maintain temporal context, and an additional forward pass to construct KV caches for subsequent generation~\cite{fime2025automatic}. These recurring operations remain a significant bottleneck to efficient long-video generation.

\begin{figure}[t]
\centering
\includegraphics[width=0.85\columnwidth]{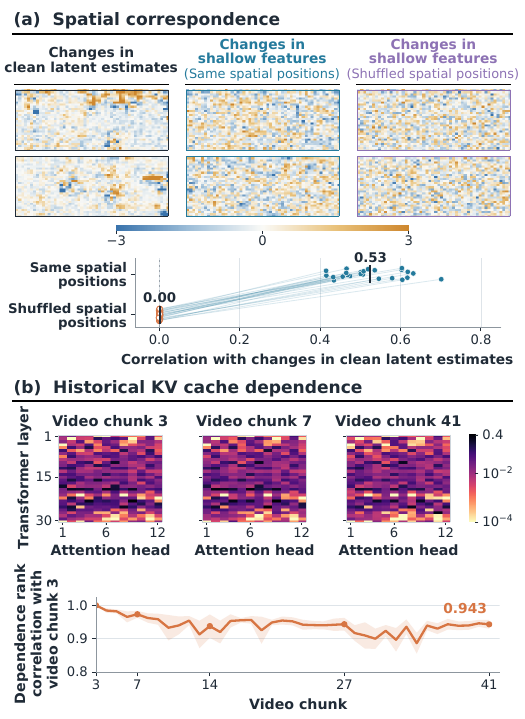}
\caption{Motivating observations for ScoutCache. (a) Changes in shallow features exhibit spatial correspondence with changes in clean latent estimates, while shuffling spatial positions disrupts this correlation. (b) Dependence on the historical KV cache varies across Transformer layers and attention heads but remains relatively stable across video chunks.}
\label{fig:intro}
\end{figure}

Existing training-free acceleration methods reduce video diffusion costs by reusing model outputs across denoising steps~\cite{liu2025timestep, lyu2025fastercache}, caching attention features~\cite{zhao2025real}, or adapting cache reuse to video content~\cite{kahatapitiya2025adaptive}. Recent studies have also explored computation reuse across video chunks~\cite{ma2026flow} and more selective access to historical KV caches~\cite{ji2026forcing, samuel2026fast}. However, these approaches face different constraints in few-step autoregressive generation. With only a few denoising steps per chunk, there are fewer opportunities to reuse previous results, while changes in the current noisy input may still affect the prediction. Moreover, reducing historical attention requires identifying which past information is useful, and compressing KV caches after their construction leaves the computation spent on discarded representations unchanged. These challenges motivate a closer examination of how denoising predictions evolve and how historical information is utilized during few-step autoregressive video generation.

Examining consecutive denoising steps reveals that useful information about prediction updates is already present in shallow features. As shown in Figure~\ref{fig:intro} (a), changes in shallow features between consecutive denoising steps exhibit spatial correspondence with changes in the clean latent estimates produced by the full denoising network.
This observation indicates that shallow features already capture part of how the prediction evolves, providing an opportunity to reduce repeated deep-network computation.
Historical attention exhibits a different pattern. As shown in Figure~\ref{fig:intro} (b), different Transformer layers and attention heads rely on the historical KV cache to different degrees, but these differences remain largely stable across consecutive video chunks. 
This stability suggests that historical attention behavior observed in earlier chunks can guide subsequent generation, reducing the need to repeatedly assess how much of the historical KV cache each layer and head requires.
Beyond reading the historical KV cache, constructing the cache for subsequent video chunks also incurs computational cost. After each chunk is generated, the model performs an additional forward pass to construct its KV representations. Compressing these representations afterward leaves the computation for discarded positions unchanged, highlighting the need to reduce the cost of KV cache construction itself.

Motivated by these findings, we propose ScoutCache, a unified training-free inference acceleration framework integrating denoising and KV cache optimization for few-step autoregressive video diffusion models. 
Specifically, ScoutCache introduces a Conditional Denoising Readout (CDR) that estimates clean latent updates from shallow feature changes and known noise variations. Calibrated on early video chunks, this readout bypasses most Transformer layers at an intermediate denoising step while retaining the final refinement by the full network. To improve historical attention efficiency, we develop Output Risk Guided Historical KV Reading (ORHR), which measures the effect of removing cached frames on attention outputs and allocates reading budgets across Transformer layers and heads. The resulting allocation is reused across subsequent video chunks to avoid repeated evaluation. 
Finally, we introduce Content and Coverage Guided Sparse Cache Construction (CCSC) that selects spatial positions based on content changes and spatial coverage before Transformer processing. This strategy reduces KV cache construction costs through sparse computation, while retaining dense representations of recent frames and a fixed historical anchor to preserve useful historical context.
All components operate without modifying pretrained model parameters.


Experiments on VBench~\cite{huang2024vbench} and MovieGen~\cite{polyak2024movie} with Self-Forcing~\cite{huang2026self} and LongLive~\cite{yang2025longlive} demonstrate improvements in both inference efficiency and video quality. For shorter videos, ScoutCache achieves 2.19$\times$ and 1.83$\times$ speedups with quality improvements of 1.52\% and 1.21\%, respectively. For longer videos, it achieves 2.28$\times$ and 1.86$\times$ speedups with corresponding quality improvements of 5.44\% and 1.11\%. Our main contributions are summarized as follows:
\begin{itemize}
    \item We highlight the importance of shallow feature changes and stable historical attention dependencies for efficient few-step autoregressive video generation.
    \item We propose ScoutCache, a training-free framework that jointly optimizes denoising, historical KV reading, and cache construction.
    \item Extensive experiments demonstrate that ScoutCache achieves state-of-the-art performance, improving both inference efficiency and video quality.
\end{itemize}

\section{Related work}
\label{sec:rw}




\noindent\textbf{Autoregressive Video Generation.}
Autoregressive video generation enables long video synthesis by generating successive frames or chunks conditioned on previously generated content. Early approaches improved temporal continuity through chunkwise conditioning and progressive denoising~\cite{henschel2025streamingt2v, xie2025progressive}, while asynchronous diffusion and continuous visual representations introduced alternative generation formulations~\cite{sun2025ar, deng2025autoregressive}. Recent studies have further explored flexible generation schedules~\cite{ma2026flex} and real-time streaming through few-step diffusion~\cite{huang2026self, yang2025longlive}, and adversarial post-training~\cite{lin2026autoregressive}. Other efforts improve causal distillation~\cite{zhu2026causal} and long-context modeling~\cite{chen2026context} to enhance generation quality and temporal consistency. Despite these advances, autoregressive generation still requires repeated denoising and historical context processing across video chunks. 
ScoutCache addresses these costs by estimating denoising updates from shallow features, allocating historical KV reading budgets based on attention output perturbations, and selecting spatial tokens before Transformer processing.

\noindent\textbf{Diffusion Acceleration.}
Diffusion acceleration aims to reduce the computational overhead of iterative denoising. Existing methods exploit temporal redundancy by reusing encoder features~\cite{ma2024deepcache}, intermediate block outputs~\cite{wimbauer2024cache}, or feature representations across denoising steps~\cite{so2024frdiff, li2023faster}. Recent approaches extend feature caching to diffusion Transformers, adapting reuse strategies according to network depth and denoising stages~\cite{chen2024delta, selvaraju2024fora}. Other studies reduce redundant cross-attention computation~\cite{zhang2024cross} or merge similar tokens to accelerate inference~\cite{bolya2023token}. However, these methods are primarily developed for conventional diffusion models, and their reuse strategies may be less effective in few-step autoregressive generation. ScoutCache addresses this challenge by estimating current denoising updates from shallow features, reducing deep-network computation while preserving final prediction refinement.

\noindent\textbf{KV Cache Optimization.}
KV cache optimization plays an important role in reducing the memory and attention costs of autoregressive video generation. Recent methods explore sparse historical attention~\cite{lv2026light} and adaptive KV selection based on frame relevance and attention head characteristics~\cite{cai2026focused, chen2026pyramid}. Other approaches estimate future cache utility~\cite{luo2026future}, identify informative historical tokens~\cite{xiao2026decoprune}, or consolidate older representations into compact memory~\cite{lu2026fademem}. KV quantization~\cite{xi2026quant} and cache maintenance strategies~\cite{li2026rolling} further improve memory efficiency during long video generation. However, these methods mainly focus on which historical representations to retain or access. ScoutCache further considers the cost of generating these representations, combining calibrated historical KV reading with sparse cache construction to reduce computation.
\section{Method}
\label{sec:method}



\subsection{Problem Formulation}
\label{sec:problem_formulation}


We consider the inference-time acceleration of pretrained
few-step autoregressive video diffusion models that generate
latent chunks conditioned on previously generated content.
Given a text condition $c$, let $x_b^{(s)}$ and $z_b^{(s)}$
denote the noisy input and predicted clean latent of chunk
$b$ at denoising step $s$, respectively. Each chunk contains
three latent frames and undergoes four denoising steps,
with re-noising applied after the first three steps:

\begin{equation}
\label{eq:ar_inference}
\begin{aligned}
z_b^{(s)} &= D_\theta(x_b^{(s)},t_s,c,\mathcal M_{b-1}),\\
x_b^{(s+1)} &= (1-\sigma_{s+1})z_b^{(s)}
+\sigma_{s+1}\varepsilon_b^{(s+1)}.
\end{aligned}
\end{equation}
Here, $D_\theta$ is the denoising network with fixed
pretrained parameters $\theta$, $\sigma_s=t_s/1000$ is the
noise coefficient, and $\varepsilon_b^{(s)}\sim\mathcal{N}(0,I)$
is Gaussian noise, with $x_b^{(1)}=\varepsilon_b^{(1)}$.
The historical KV cache $M_{b-1}$ provides temporal context from
preceding chunks. After denoising, $z_b^{(4)}$ is decoded
into video frames, and an additional forward pass at the
zero-noise timestep constructs KV representations for
subsequent chunks.



\begin{figure*}[t]
\centering
\includegraphics[width=0.85\textwidth]{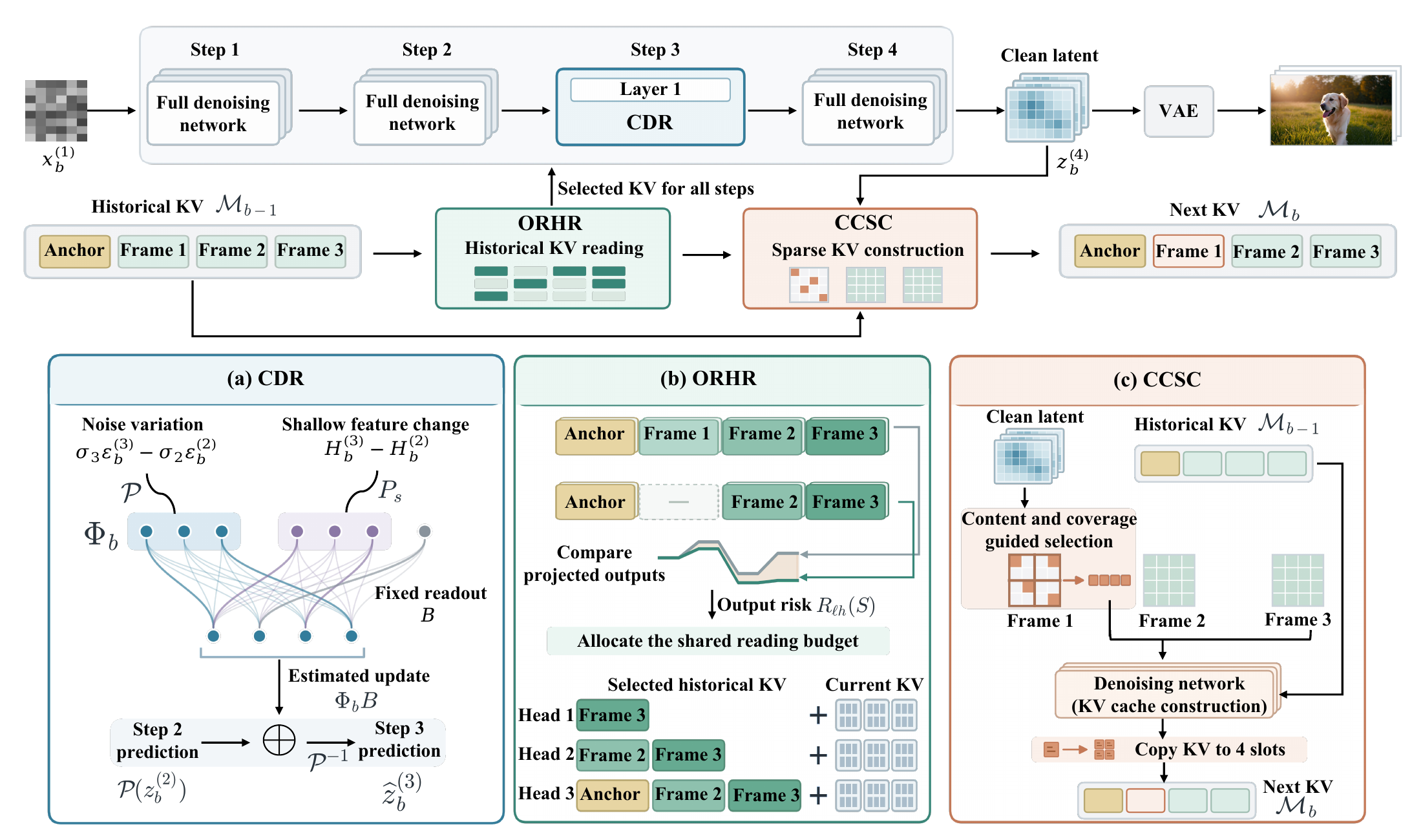}
\caption{The top shows the overall inference pipeline, integrating denoising acceleration and KV cache optimization. The bottom illustrates three components. (a) CDR estimates clean latent updates from noise variations and shallow feature changes. (b) ORHR allocates historical KV reading budgets based on attention output perturbations. (c) CCSC reduces cache construction costs through content and coverage guided spatial token selection.}
\label{fig:model}
\end{figure*}

As illustrated in Figure~\ref{fig:model}, ScoutCache integrates three complementary components. CDR estimates intermediate clean latent updates from shallow feature changes and known noise variations.
ORHR allocates historical KV reading budgets across layers
and heads according to attention output perturbations,
while CCSC reduces cache construction costs by selecting
spatial tokens before Transformer processing. CDR and ORHR
are calibrated on early chunks, and all components operate
without modifying pretrained model parameters.

\subsection{Conditional Denoising Readout}
\label{sec:cdr}

Motivated by the spatial correspondence observed in Figure~\ref{fig:intro} (a), CDR estimates clean latent updates from shallow feature changes and known noise variations. As illustrated in Figure~\ref{fig:model} (a), it uses the full prediction at the second denoising step as a reference, evaluates only the first Transformer layer at the third step to estimate the update, and retains the full fourth step for final refinement.
Let $H_b^{(2)},H_b^{(3)}\in\mathbb R^{N\times d}$ denote the first Transformer layer outputs at the second and third denoising steps, where $N$ is the number of tokens per chunk and $d$ is the hidden dimension. To characterize the prediction update, we construct a conditioning feature that combines the known noise variation with the corresponding shallow feature difference:
\begin{equation}
\label{eq:cdr_feature}
\Phi_b=\Big[
\mathcal P\big(\sigma_3\varepsilon_b^{(3)}
-\sigma_2\varepsilon_b^{(2)}\big),
(H_b^{(3)}-H_b^{(2)})P_s,\mathbf 1_N
\Big],
\end{equation}
where $\mathcal P$ rearranges the latent into $N$ patches of
dimension $p$, and $P_s\in\mathbb R^{d\times r_s}$ is a fixed
Gaussian projection with entries independently sampled from
$\mathcal N(0,1/d)$, and $r_s$ denotes the shallow feature
projection dimension. The resulting $\Phi_b\in\mathbb R^{N\times m}$, with $m=p+r_s+1$, includes
a constant column $\mathbf 1_N$ for the intercept.
The noise term accounts for stochastic variations introduced
by sampling, while the shallow feature difference captures
the network response to the current latent and historical
context. We calibrate the mapping from $\Phi_b$ to clean latent updates using the second and third generated chunks, where full denoising predictions are available.
For each calibration chunk, the target update is defined as $Y_b=\mathcal P(z_b^{(3)}-z_b^{(2)})$. Stacking the conditioning features and target updates from both chunks yields $X\in\mathbb R^{2N\times m}$ and $Y\in\mathbb R^{2N\times p}$. To account for differences in feature scales, we normalize the columns of $X$ using a diagonal RMS scaling matrix $D$, whose diagonal entries are lower bounded by $10^{-3}$. The readout mapping is then obtained through regularized least squares:
\begin{equation}
\label{eq:cdr_mapping}
\begin{aligned}
Z &= XD^{-1},\\
B &= D^{-1}(Z^\top Z+\lambda\Lambda)^{-1}Z^\top Y.
\end{aligned}
\end{equation}
where $Z\in\mathbb R^{2N\times m}$ is the normalized conditioning matrix, and $B\in\mathbb R^{m\times p}$ maps conditioning features to clean latent updates.
$\Lambda=\operatorname{diag}(1,\ldots,1,0)$ excludes the intercept from regularization, and $\lambda=m$ controls the regularization strength. Feature normalization balances the contributions of noise variations and shallow feature differences, while regularization stabilizes the mapping estimated from a short video prefix. The resulting $B$ is computed once per video and shared across subsequent chunks.

For subsequent chunks $b\geq4$, CDR executes only the first Transformer layer at the third denoising step and estimates the clean latent as:
\begin{equation}
\widehat z_b^{(3)}
=\mathcal P^{-1}\left(
\mathcal P(z_b^{(2)})+\Phi_bB
\right),
\end{equation}
where $\widehat z_b^{(3)}$ denotes the estimated clean latent at the third denoising step, and $\mathcal P^{-1}$ restores the original latent layout. The predicted update $\Phi_bB$ adapts to the current chunk, allowing CDR to bypass the remaining Transformer layers at this step. The estimated latent is then used to construct the next noisy input following Eq.~\ref{eq:ar_inference}, while the fourth denoising step is evaluated using the full network for final refinement.

\subsection{Output Risk Guided Historical KV Reading}
The varying yet stable historical dependencies observed in Figure~\ref{fig:intro} (b) motivate ORHR to allocate historical reading budgets based on the impact of cache removal on attention outputs, as illustrated in Figure~\ref{fig:model} (b). The historical cache consists of four frame regions $\mathcal R=\{0,1,2,3\}$, including the first latent frame of the video as a fixed anchor and the three frames from the preceding chunk. The current chunk, denoted by $\ast$, always participates in attention. For each Transformer layer $\ell$ and attention head $h$, ORHR selects a nonempty subset $S_{\ell h}\subseteq\mathcal R$ of historical regions to access.

To efficiently evaluate different historical subsets, we decompose attention into reusable regional statistics. For a query $q$, let $o_r(q)$ denote the attention output normalized within region $r$, and $\eta_r(q)=\log\sum_{j\in r}\exp(q^\top k_j/\sqrt{d_h})$ denote its log normalization factor, where $k_j$ is the key vector at position $j$ and $d_h$ is the attention head dimension. Here, $o_r(q)$ is the attention weighted average of the corresponding value vectors. The attention output for any historical subset $S\subseteq\mathcal R$ can be reconstructed as:
\begin{equation}
\label{eq:orhr_attention}
a(q;S)=
\frac{
\sum_{r\in S\cup\{\ast\}}
\exp(\eta_r(q))o_r(q)
}{
\sum_{r\in S\cup\{\ast\}}
\exp(\eta_r(q))
},
\end{equation}
where $a(q;S)$ denotes the attention output using the selected historical regions and the current chunk. This formulation preserves softmax normalization while allowing regional statistics to be reused across candidate subsets. ORHR evaluates the perturbation after attention output projection using regional statistics collected from the third generated chunk over a fixed query set $\mathcal Q$. Let $\mathcal T_\ell$ denote the calibration steps for layer $\ell$, including all four steps for the first layer and steps 1, 2, and 4 for deeper layers, consistent with CDR. The output risk of a historical subset $S$ is defined as:
\begin{equation}
\label{eq:orhr_risk}
\begin{aligned}
e_{\ell h}^{(s)}(q;S)
&=\frac{[a_{\ell h}^{(s)}(q;S)-a_{\ell h}^{(s)}(q;\mathcal R)]W_{\ell h}^{O}\Pi}{\rho_{\ell s}},\\
R_{\ell h}(S)
&=\frac{
\sum_{s\in\mathcal T_\ell}\sum_{q\in\mathcal Q}
\|e_{\ell h}^{(s)}(q;S)\|_2^2
}{
|\mathcal T_\ell||\mathcal Q|r_o
},
\end{aligned}
\end{equation}
where $a_{\ell h}^{(s)}$ denotes the attention output at layer $\ell$, head $h$, and step $s$, and $W_{\ell h}^{O}\in\mathbb R^{d_h\times d}$ is the corresponding output projection. $\Pi\in\mathbb R^{d\times r_o}$ is a fixed Gaussian projection with entries independently sampled from $\mathcal N(0,1/r_o)$, where $r_o$ denotes the projection dimension. The normalization factor $\rho_{\ell s}$ is the RMS magnitude of the projected full-history attention output aggregated across heads and measured over calibration queries and channels, lower bounded by $10^{-4}$. Accordingly, $e_{\ell h}^{(s)}(q;S)$ represents the normalized output perturbation, and $R_{\ell h}(S)$ measures its average squared magnitude, ensuring comparable risk scales across layers and steps.

Based on the estimated risks, we allocate a global historical reading budget across the network. Since the squared norm of the aggregated head perturbations is bounded by $H$ times the sum of their squared norms, minimizing individual risks provides a tractable surrogate for controlling the combined output perturbation. We formulate the allocation as:
\begin{equation}
\label{eq:orhr_allocation}
\begin{aligned}
\min_{\{S_{\ell h}\}}\quad&
\sum_{\ell=1}^{L}\sum_{h=1}^{H}
R_{\ell h}(S_{\ell h}),\\
\text{s.t.}\quad&
\sum_{\ell,h}|S_{\ell h}|\leq\mathcal B,
\quad 1\leq|S_{\ell h}|\leq4,
\end{aligned}
\end{equation}
where $L$ and $H$ denote the numbers of Transformer layers and attention heads, $|S_{\ell h}|$ is the number of selected historical frames, and $\mathcal B$ is the global reading budget in frame accesses. We solve Eq.~\ref{eq:orhr_allocation} greedily, initializing each head with its lowest risk historical frame and iteratively adding the frame with the largest marginal risk reduction. We set $\mathcal B=2LH$, providing an average budget of two historical frames per head while allowing individual heads to access one to four frames. The resulting policy is calibrated once and reused across subsequent chunks as recent regions are updated. During inference, selected historical KV representations are packed with current chunk representations for variable length attention computation.

\subsection{Content and Coverage Guided Sparse Cache Construction}
While ORHR reduces historical KV reading costs, CCSC targets the additional forward pass for KV cache construction by selecting informative spatial positions before Transformer processing. The first three video chunks use dense cache construction for calibration. From the fourth chunk onward, CCSC constructs sparse representations for the first frame while retaining dense representations for the remaining two frames, as illustrated in Figure~\ref{fig:model} (c).

For each generated latent frame $f$, we apply $2\times2$ average pooling to its clean latent prediction to obtain spatial proxy features $g_f(i)\in\mathbb R^{C_z}$ aligned with the Transformer token grid, where $i$ denotes a spatial position and $C_z$ is the latent channel dimension. Each position maintains its last selected proxy $\bar g(i)$, a visited indicator $v(i)$, and an age $a_f(i)$ counting frames since its last selection. We define the content change and selection priority as:
\begin{equation}
\label{eq:ccsc_priority}
\begin{aligned}
\delta_f(i)
&=\frac{\|g_f(i)-\bar g(i)\|_2}{\sqrt{C_z}},\\
u_f(i)
&=\delta_f(i)\\
&\quad +(1+\max_j\delta_f(j))
\mathbf{1}[v(i)=0\lor a_f(i)\ge A],
\end{aligned}
\end{equation}
where $\delta_f(i)$ measures the content change since the position was last selected, and $A$ is the staleness threshold. The stored proxies are initialized using the spatial mean of the first latent frame. For unvisited or stale positions, the additional priority term exceeds the maximum content change in the current frame, promoting their refresh. 
To maintain spatial coverage, we divide the token grid into nonoverlapping $2\times2$ groups $\mathcal G$ and select the highest priority position in each group as $\mathcal I_f=\{\arg\max_{i\in G}u_f(i)\mid G\in\mathcal G\}$, where $G$ denotes a spatial group and $\mathcal I_f$ contains $P/4$ selected positions.
For each selected position $i\in\mathcal I_f$, we update $\bar g(i)=g_f(i)$, set $v(i)=1$, and reset $a_f(i)=0$, while the ages of unselected positions increase by one per frame. These states are updated sequentially across generated frames, while sparse cache construction is applied to the first frame of each chunk from the fourth chunk onward.
For each chunk $b\geq4$, the positions processed during cache construction are:
\begin{equation}
\label{eq:ccsc_tokens}
\mathcal J_b=
(\{1\}\times\mathcal I_{b,1})
\cup
(\{2,3\}\times\{1,\ldots,P\}),
\end{equation}
where $\mathcal I_{b,1}$ denotes the selected positions in the first frame of chunk $b$, $P$ is the number of spatial tokens per latent frame, and the indices $\{1,2,3\}$ identify its three latent frames. Consequently, $|\mathcal J_b|=P/4+2P=3N/4$, reducing the number of processed tokens by 25\% while preserving dense representations for the two most recent frames.

\begin{table*}[t]
\centering
\caption{Quantitative comparison on Self-Forcing across different video durations. Latency is measured in seconds. \textcolor{blue!70!black}{$\blacktriangle$} indicates a Total Score higher than the original backbone. The best and second best results are highlighted in \textbf{bold} and \underline{underline}, respectively.}
\label{tab:self_forcing}
\resizebox{\textwidth}{!}{
\begin{tabular}{l*{16}{c}}
\toprule
Method
& \multicolumn{4}{c}{VBench (5s)}
& \multicolumn{4}{c}{MovieGen (30s)}
& \multicolumn{4}{c}{MovieGen (60s)}
& \multicolumn{4}{c}{MovieGen (120s)} \\
\cmidrule(lr){2-5}
\cmidrule(lr){6-9}
\cmidrule(lr){10-13}
\cmidrule(lr){14-17}
& Latency$\downarrow$ & FPS$\uparrow$ & Speedup$\uparrow$ & Total$\uparrow$
& Latency$\downarrow$ & FPS$\uparrow$ & Speedup$\uparrow$ & Total$\uparrow$
& Latency$\downarrow$ & FPS$\uparrow$ & Speedup$\uparrow$ & Total$\uparrow$
& Latency$\downarrow$ & FPS$\uparrow$ & Speedup$\uparrow$ & Total$\uparrow$ \\
\midrule

Self-Forcing (NeurIPS'25)
& 9.00 & 8.89 & 1.00$\times$ & 81.52
& 65.59 & 7.32 & 1.00$\times$ & 79.10
& 131.90 & 7.28 & 1.00$\times$ & 77.76
& 264.32 & 7.26 & 1.00$\times$ & 75.03 \\

FlowCache (ICLR'26)
& 7.22 & 11.08 & 1.25$\times$ & 81.56\scoreup
& 52.82 & 9.09 & 1.24$\times$ & 78.45
& 105.71 & 9.08 & 1.25$\times$ & 76.60
& 212.24 & 9.05 & 1.25$\times$ & 74.62 \\

TeaCache (CVPR'25)
& 7.39 & 10.83 & 1.22$\times$ & 81.07
& 54.05 & 8.88 & 1.21$\times$ & 78.54
& 109.33 & 8.78 & 1.21$\times$ & 77.20
& 218.52 & 8.79 & 1.21$\times$ & 74.77 \\

MotionCache-Slow (ICML'26)
& 7.53 & 10.62 & 1.20$\times$ & 68.92
& 55.35 & 8.67 & 1.19$\times$ & 68.97
& 111.68 & 8.60 & 1.18$\times$ & 68.71
& 223.17 & 8.60 & 1.18$\times$ & 68.56 \\

MotionCache-Fast (ICML'26)
& 6.54 & 12.24 & 1.38$\times$ & 66.86
& 46.76 & 10.27 & 1.40$\times$ & 68.72
& 94.80 & 10.13 & 1.39$\times$ & 68.57
& 186.26 & 10.31 & 1.42$\times$ & 68.91 \\

SVG2 (NeurIPS'25)
& 11.61 & 6.89 & 0.78$\times$ & 80.87
& 77.15 & 6.22 & 0.85$\times$ & 77.35
& 155.46 & 6.18 & 0.85$\times$ & 75.77
& 308.24 & 6.23 & 0.86$\times$ & 74.25 \\

Light Forcing (ICML'26)
& 6.65 & 12.03 & 1.35$\times$ & 81.08
& 42.79 & 11.22 & 1.53$\times$ & 78.47
& 85.91 & 11.17 & 1.54$\times$ & 78.03\scoreup
& 170.01 & 11.29 & 1.55$\times$ & 77.62\scoreup \\

Dummy Forcing (arXiv'26)
& \underline{6.48} & \underline{12.35} & \underline{1.39$\times$} & 81.04
& \underline{38.58} & \underline{12.44} & \underline{1.70$\times$} & \underline{80.11}\scoreup
& \underline{76.73} & \underline{12.51} & \underline{1.72$\times$} & \underline{79.51}\scoreup
& \underline{152.58} & \underline{12.58} & \underline{1.73$\times$} & \underline{79.02}\scoreup \\

Forcing-KV (arXiv'26)
& 6.51 & 12.29 & 1.38$\times$ & \underline{81.65}\scoreup
& 40.43 & 11.87 & 1.62$\times$ & 79.76\scoreup
& 80.03 & 12.00 & 1.65$\times$ & 79.26\scoreup
& 159.73 & 12.02 & 1.65$\times$ & 78.70\scoreup \\

\midrule
\textbf{ScoutCache (Ours)}
& \textbf{5.64} & \textbf{14.18} & \textbf{1.60$\times$} & \textbf{81.76}\scoreup
& \textbf{29.99} & \textbf{16.01} & \textbf{2.19$\times$} & \textbf{80.30}\scoreup
& \textbf{58.65} & \textbf{16.37} & \textbf{2.25$\times$} & \textbf{79.87}\scoreup
& \textbf{115.78} & \textbf{16.58} & \textbf{2.28$\times$} & \textbf{79.11}\scoreup \\
\bottomrule
\end{tabular}
}
\end{table*}

\begin{table*}[t]
\centering
\caption{Quantitative comparison on LongLive across different video durations. Latency is measured in seconds. \textcolor{blue!70!black}{$\blacktriangle$} indicates a Total Score higher than the original backbone. The best and second best results are highlighted in \textbf{bold} and \underline{underline}, respectively.}
\label{tab:longlive}
\resizebox{\textwidth}{!}{
\begin{tabular}{l*{16}{c}}
\toprule
Method
& \multicolumn{4}{c}{VBench (5s)}
& \multicolumn{4}{c}{MovieGen (30s)}
& \multicolumn{4}{c}{MovieGen (60s)}
& \multicolumn{4}{c}{MovieGen (120s)} \\
\cmidrule(lr){2-5}
\cmidrule(lr){6-9}
\cmidrule(lr){10-13}
\cmidrule(lr){14-17}
& Latency$\downarrow$ & FPS$\uparrow$ & Speedup$\uparrow$ & Total$\uparrow$
& Latency$\downarrow$ & FPS$\uparrow$ & Speedup$\uparrow$ & Total$\uparrow$
& Latency$\downarrow$ & FPS$\uparrow$ & Speedup$\uparrow$ & Total$\uparrow$
& Latency$\downarrow$ & FPS$\uparrow$ & Speedup$\uparrow$ & Total$\uparrow$ \\
\midrule

LongLive (ICLR'26)
& 8.58 & 9.32 & 1.00$\times$ & 79.63
& 54.77 & 8.76 & 1.00$\times$ & 80.35
& 108.38 & 8.86 & 1.00$\times$ & 80.04
& 215.83 & 8.90 & 1.00$\times$ & 80.06 \\

FlowCache (ICLR'26)
& 7.66 & 10.44 & 1.12$\times$ & 79.70\scoreup
& 48.59 & 9.88 & 1.13$\times$ & 80.09
& 96.64 & 9.93 & 1.12$\times$ & 79.99
& 193.45 & 9.92 & 1.12$\times$ & 79.88 \\

TeaCache (CVPR'25)
& 6.91 & 11.57 & 1.24$\times$ & \underline{80.14}\scoreup
& 44.07 & 10.89 & 1.24$\times$ & 79.89
& 87.95 & 10.91 & 1.23$\times$ & 79.58
& 175.25 & 10.96 & 1.23$\times$ & 79.68 \\

SVG2 (NeurIPS'25)
& 11.14 & 7.18 & 0.77$\times$ & 79.79\scoreup
& 66.57 & 7.21 & 0.82$\times$ & 79.37
& 132.91 & 7.22 & 0.82$\times$ & 79.37
& 266.31 & 7.21 & 0.81$\times$ & 79.42 \\

Dummy Forcing (arXiv'26)
& \underline{6.43} & \underline{12.44} & \underline{1.33$\times$} & 79.50
& \underline{37.70} & \underline{12.73} & \underline{1.45$\times$} & 79.64
& \underline{74.98} & \underline{12.80} & \underline{1.45$\times$} & 79.43
& \underline{149.04} & \underline{12.88} & \underline{1.45$\times$} & 79.30 \\

Forcing-KV (arXiv'26)
& 6.50 & 12.30 & 1.32$\times$ & \textbf{80.39}\scoreup
& 40.61 & 11.82 & 1.35$\times$ & \underline{80.78}\scoreup
& 81.17 & 11.83 & 1.34$\times$ & \underline{80.52}\scoreup
& 161.61 & 11.88 & 1.34$\times$ & \underline{80.44}\scoreup \\

\midrule
\textbf{ScoutCache (Ours)}
& \textbf{5.64} & \textbf{14.18} & \textbf{1.52$\times$} & 79.85\scoreup
& \textbf{29.96} & \textbf{16.02} & \textbf{1.83$\times$} & \textbf{81.32}\scoreup
& \textbf{58.50} & \textbf{16.41} & \textbf{1.85$\times$} & \textbf{81.25}\scoreup
& \textbf{115.78} & \textbf{16.58} & \textbf{1.86$\times$} & \textbf{80.95}\scoreup \\
\bottomrule
\end{tabular}
}
\end{table*}

After patch embedding, only the selected tokens enter the Transformer, with their original spatiotemporal coordinates preserved for rotary positional encoding. However, representing four neighboring positions with a single token reduces their aggregate contribution to attention normalization. To compensate for this effect, we introduce multiplicity weighted attention:
\begin{equation}
\label{eq:ccsc_attention}
\operatorname{Attn}(q)=
\operatorname{softmax}\left(
\frac{qK^\top}{\sqrt{d_h}}
+\log\mathbf m^\top
\right)V.
\end{equation}
where $q$ is the query vector, $K$ and $V$ contain the participating key and value vectors, and $d_h$ is the attention head dimension. The multiplicity vector $\mathbf m$ assigns $m_j=4$ to sparse representatives and $m_j=1$ to dense tokens, accounting for the four spatial slots represented by each sparse token in attention normalization. We implement this weighting by replicating each representative's position-encoded KV vectors into its corresponding four slots, preserving the regular cache layout without additional Transformer computation.
The new KV representations are constructed using the existing historical cache and replace the previous recent chunk only after all Transformer layers have been processed. The fixed dense anchor remains unchanged, while the expanded KV representations form the updated three-frame history, maintaining compatibility with ORHR's historical reading policy.

\section{Experiments}
\label{sec:exper}

\subsection{Experimental Setup}
\noindent\textbf{Implementation Details.} All experiments are conducted on a single NVIDIA A100 40GB GPU using PyTorch with BF16 precision and a batch size of 1. Videos are generated at 832 × 480 resolution and 16 FPS, with four denoising steps per chunk of three latent frames. ScoutCache performs full computation for the first three chunks and enables acceleration thereafter. CDR uses a shallow feature projection dimension $r_s=128$ and regularization strength $\lambda=m=193$, with the readout mapping calibrated on chunks 2–3 and applied at the third denoising step. ORHR uses a random projection dimension $r_o=32$, 120 calibration queries from the third chunk, and a global historical reading budget $\mathcal B=720=2LH$. CCSC sets the staleness threshold $A=18$ and selects one position per $2\times2$ spatial group, reducing cache construction to 3,510 processed tokens per chunk.

\noindent\textbf{Backbones and Baselines.} We evaluate ScoutCache on Self-Forcing~\cite{huang2026self} and LongLive~\cite{yang2025longlive}, using their publicly released 1.3B checkpoints. We compare with six training-free methods, including TeaCache~\cite{liu2025timestep}, FlowCache~\cite{ma2026flow}, MotionCache~\cite{xu2026motion}, SVG2~\cite{yang2026sparse}, Dummy Forcing~\cite{guo2026efficient}, and Forcing-KV~\cite{ji2026forcing}, 
as well as Light Forcing~\cite{lv2026light}, which requires 2,000 additional post-training iterations on pretrained weights. We evaluate it using its officially released checkpoint.
The original backbones serve as references for both quality and efficiency. 
For all training-free baselines, we adopt their official implementations and released hyperparameter configurations, including method-specific caching settings. Within each backbone, these methods share the same pretrained weights, prompts, random seeds, resolution, and sampling schedule.

\noindent\textbf{Benchmarks and Metrics.} We evaluate video quality on VBench~\cite{huang2024vbench} and MovieGen~\cite{polyak2024movie} using five random seeds per prompt. For VBench, we generate 5-second videos from all 946 prompts and report the official Total Scores across 16 dimensions. For MovieGen, we use the first 128 prompts from Self-Forcing's extended prompt list to generate videos of 30, 60, and 120 seconds. Following Forcing-KV~\cite{ji2026forcing}, we evaluate six VBench-Long dimensions: subject consistency, background consistency, motion smoothness, dynamic degree, aesthetic quality, and imaging quality. These six dimensions are aggregated into a Total Score following the official VBench normalization rules and corresponding weights. Efficiency is measured by latency, generation FPS, and speedup. 
Latency measures the complete diffusion generation time, including online calibration, denoising, sampling updates, and cache maintenance. Speedup is the ratio of backbone to method latency, while generation FPS denotes inference throughput.

\subsection{Main Results}
\begin{table}[t]
\centering
\caption{Ablation studies of ScoutCache on MovieGen. Additional evaluations verify the component contributions across video durations and backbones. Latency is measured in seconds.}
\label{tab:ablation}
\footnotesize
\setlength{\tabcolsep}{3pt}
\begin{tabular}{lccc}
\toprule
Configuration & Latency$\downarrow$ & Speedup$\uparrow$ & Total$\uparrow$ \\
\midrule

\multicolumn{4}{l}{\textit{Main Ablation: Self-Forcing, 30s}} \\
Original Backbone & 65.59 & 1.00$\times$ & 79.10 \\
Fixed History (w/o All) & 39.47 & 1.66$\times$ & 80.09 \\
CDR Only & 32.23 & 2.04$\times$ & 79.82 \\
w/o CDR & 36.29 & 1.81$\times$ & 80.22 \\
w/o ORHR & 30.58 & 2.15$\times$ & 79.83 \\
w/o CCSC & 30.93 & 2.12$\times$ & 80.09 \\
w/ Fixed 2 Frames/Head & 30.10 & 2.18$\times$ & 79.91 \\
w/ Preprojection Risk & 30.06 & 2.18$\times$ & 80.03 \\
ScoutCache (Full)
& \textbf{29.99} & \textbf{2.19$\times$} & \textbf{80.30} \\

\midrule
\multicolumn{4}{l}{\textit{Cross Duration: Self-Forcing, 120s}} \\
w/o CDR & 144.15 & 1.83$\times$ & 78.76 \\
w/o ORHR & 118.58 & 2.23$\times$ & 78.66 \\
w/o CCSC & 123.10 & 2.15$\times$ & 79.02 \\
ScoutCache (Full)
& \textbf{115.78} & \textbf{2.28$\times$} & \textbf{79.11} \\

\midrule
\multicolumn{4}{l}{\textit{Cross Backbone: LongLive, 30s}} \\
w/o CDR & 36.23 & 1.51$\times$ & 80.91 \\
w/o ORHR & 31.12 & 1.76$\times$ & 80.61 \\
w/o CCSC & 31.09 & 1.76$\times$ & 81.09 \\
ScoutCache (Full)
& \textbf{29.96} & \textbf{1.83$\times$} & \textbf{81.32} \\
\bottomrule
\end{tabular}
\end{table}

\begin{figure*}[t]
\centering
\includegraphics[width=0.85\textwidth]{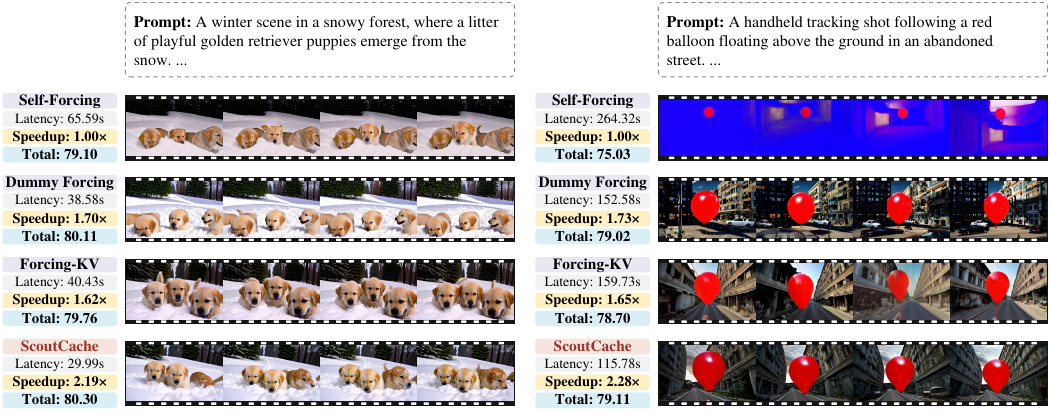}
\caption{Qualitative comparison of video generation results. We show four sampled frames for each prompt, comparing ScoutCache with the original backbones, Dummy Forcing, and Forcing-KV.}
\label{fig:qualitative-self}
\end{figure*}

\begin{figure}[t]
\centering
\includegraphics[width=0.9\columnwidth]{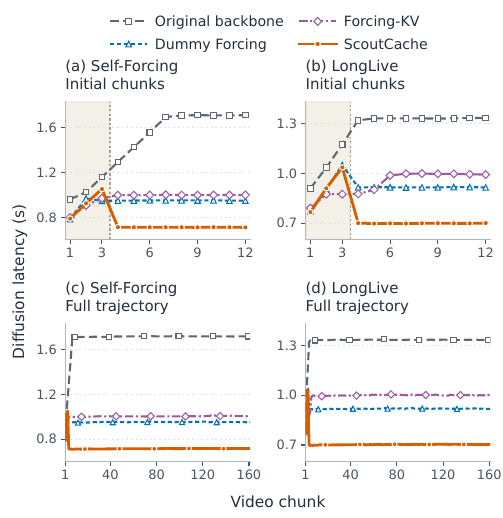}
\caption{Chunkwise diffusion latency on Self-Forcing and LongLive. (a,b) Latency over the first 12 chunks, showing calibration and subsequent acceleration. (c,d) Latency across 160 chunks.}
\label{fig:rb1}
\end{figure}

\noindent\textbf{Results on Self-Forcing.}
Table~\ref{tab:self_forcing} compares ScoutCache with existing acceleration methods on Self-Forcing across different video durations. ScoutCache achieves the lowest latency, highest generation FPS, and highest Total Score in all four settings. Specifically, it delivers a 1.60$\times$ speedup on VBench and 2.19$\times$, 2.25$\times$, and 2.28$\times$ speedups on MovieGen at 30, 60, and 120 seconds, respectively. The increasing speedup with video duration reflects the reduced relative overhead of initial calibration. Notably, for 120-second generation, ScoutCache reduces latency from 264.32s to 115.78s while improving the Total Score from 75.03 to 79.11, corresponding to a 5.44\% relative quality improvement. We further conduct two-sided paired t-tests over five matched random seeds against the strongest competing method at each duration. The mean Total Score improvements are 0.11, 0.19, 0.36, and 0.09 points for 5s, 30s, 60s, and 120s, respectively, with corresponding 95\% confidence intervals of [0.085, 0.135], [0.165, 0.215], [0.335, 0.385], and [0.065, 0.115]. All improvements are statistically significant ($p<0.001$). 


\noindent\textbf{Results on LongLive.}
To further validate the effectiveness of ScoutCache across different autoregressive backbones, we evaluate it on LongLive. As shown in Table~\ref{tab:longlive}, ScoutCache consistently achieves the lowest latency and highest generation FPS across all video durations, delivering a 1.52$\times$ speedup on VBench and 1.83$\times$, 1.85$\times$, and 1.86$\times$ speedups on MovieGen at 30, 60, and 120 seconds, respectively. In terms of generation quality, ScoutCache improves the Total Score over the original LongLive in all settings and achieves the highest scores across all three MovieGen durations. Notably, for 120-second generation, it reduces latency from 215.83s to 115.78s while improving the Total Score from 80.06 to 80.95. We further conduct two-sided paired t-tests over five matched random seeds against Forcing-KV, the strongest competing method across all three MovieGen durations. The mean Total Score improvements are 0.54, 0.73, and 0.51 points for 30s, 60s, and 120s, respectively, with corresponding 95\% confidence intervals of [0.410, 0.670], [0.580, 0.880], and [0.348, 0.672]. All improvements are statistically significant ($p<0.001$). Together with the results on Self-Forcing, these findings demonstrate that ScoutCache consistently improves inference efficiency and generation quality across different autoregressive video diffusion backbones.

\subsection{Ablation Studies}

Table~\ref{tab:ablation} evaluates the contributions of CDR, ORHR, and CCSC, with additional experiments across video durations and backbones. On Self-Forcing, using a fixed four-frame history reduces the 30s latency from 65.59s to 39.47s, while CDR alone further reduces it to 32.23s. Combining all three components achieves 29.99s with a Total Score of 80.30. Removing CDR increases latency to 36.29s, confirming its major contribution to denoising acceleration. ORHR is important for generation quality, as its removal decreases the Total Score by 0.47 points. Assigning two historical frames to each head or estimating risk before output projection also leads to lower quality, supporting the proposed risk guided allocation. CCSC provides additional efficiency gains, with its removal increasing the 120s latency from 115.78s to 123.10s. The consistent trends on LongLive further demonstrate the complementary contributions of the three components across different backbones and video durations.



\subsection{Chunkwise Latency Analysis}
We further examine how the efficiency gains of ScoutCache evolve across video chunks. As shown in Fig.~\ref{fig:rb1}, ScoutCache achieves a substantial latency reduction from the fourth chunk onward and maintains stable performance throughout 160 chunks. After the initial calibration, its average per-chunk latency reaches 0.72s on Self-Forcing and 0.70s on LongLive, representing reductions of 25.0\% and 23.5\% compared with Dummy Forcing, respectively. While the original backbones exhibit increasing latency during early generation, ScoutCache maintains consistently low latency across subsequent chunks. These results demonstrate that ScoutCache provides sustained acceleration without increasing per-chunk overhead as video generation progresses.

\subsection{Qualitative Results}

As shown in Figure~\ref{fig:qualitative-self}, ScoutCache preserves visual details and scene structure while substantially reducing generation time. In the abandoned street example, the original Self-Forcing exhibits severe color distortions and loses much of the scene detail, while Forcing-KV shows noticeable blurring in some frames. In contrast, ScoutCache maintains clear balloon contours and recognizable building structures throughout the sampled frames. 
These observations complement the quantitative results, demonstrating that ScoutCache achieves substantial acceleration while retaining strong visual quality across diverse scenes and video durations.

Due to space limitations, further analyses of \textbf{hyperparameter sensitivity}, \textbf{runtime breakdowns}, and \textbf{component mechanisms}, along with \textbf{additional qualitative comparisons and detailed video quality scores}, are provided in the supplementary material.

\section{Conclusion}
\label{sec:conclusion}
In this paper, we introduced ScoutCache, a training-free framework for accelerating few-step autoregressive video diffusion. ScoutCache improves inference efficiency by jointly reducing the computational costs of denoising, historical KV reading, and cache construction. Specifically, we develop CDR to estimate intermediate denoising updates from shallow feature changes and known noise variations. We further introduce ORHR to allocate historical KV reading budgets based on attention output perturbations, and CCSC to reduce cache construction costs through spatial token selection guided by content changes and spatial coverage. Extensive experiments on Self-Forcing and LongLive demonstrate that ScoutCache achieves up to 2.28× and 1.86× speedups, respectively, while consistently improving generation quality across different video durations. These results demonstrate the effectiveness of jointly optimizing denoising and KV cache operations for efficient long video generation without modifying pretrained model parameters.

{\small
 \bibliographystyle{ieeenat_fullname}
 \bibliography{main}
}
\clearpage

\appendix


\section{Baseline Methods}
We compare ScoutCache with six representative training-free acceleration methods, including \textbf{TeaCache}~\cite{liu2025timestep}, \textbf{FlowCache}~\cite{ma2026flow}, \textbf{MotionCache}~\cite{xu2026motion}, \textbf{SVG2}~\cite{yang2026sparse}, \textbf{Dummy Forcing}~\cite{guo2026efficient}, and \textbf{Forcing-KV}~\cite{ji2026forcing}, as well as the training-based \textbf{Light Forcing}~\cite{lv2026light}. A brief introduction to each baseline is provided below:
\begin{itemize}
    \item \textbf{TeaCache}~\cite{liu2025timestep}: It focuses on reusing denoising outputs by estimating their changes from timestep embedding modulated inputs, with polynomial rescaling to improve estimation accuracy.

    \item \textbf{FlowCache}~\cite{ma2026flow}: To accommodate varying redundancy across video chunks, it adopts adaptive chunkwise feature caching and compresses historical KV caches based on importance and redundancy.

    \item \textbf{MotionCache}~\cite{xu2026motion}: It exploits motion differences between video frames to guide token level cache reuse, allowing more frequent updates in dynamic regions while reducing computation in static areas.

    \item \textbf{SVG2}~\cite{yang2026sparse}: By clustering and reordering tokens according to semantic similarity, it improves sparse attention efficiency through more accurate token selection and dynamic computation budgets.

    \item \textbf{Dummy Forcing}~\cite{guo2026efficient}: It exploits the varying historical context requirements of attention heads through heterogeneous memory allocation, dynamic head programming, and context packing.

    \item \textbf{Forcing-KV}~\cite{ji2026forcing}: To reduce historical KV redundancy, it distinguishes static and dynamic attention heads, applying structured pruning and similarity based pruning to their respective caches.

    \item \textbf{Light Forcing}~\cite{lv2026light}: It targets attention overhead in autoregressive video generation through chunk aware sparsity allocation and hierarchical selection of historical frames and spatial blocks. Unlike the other baselines, it requires additional post training.
\end{itemize}

\section{Hyperparameter Analysis}

\begin{figure*}[t]
\centering
\includegraphics[width=0.8\textwidth]{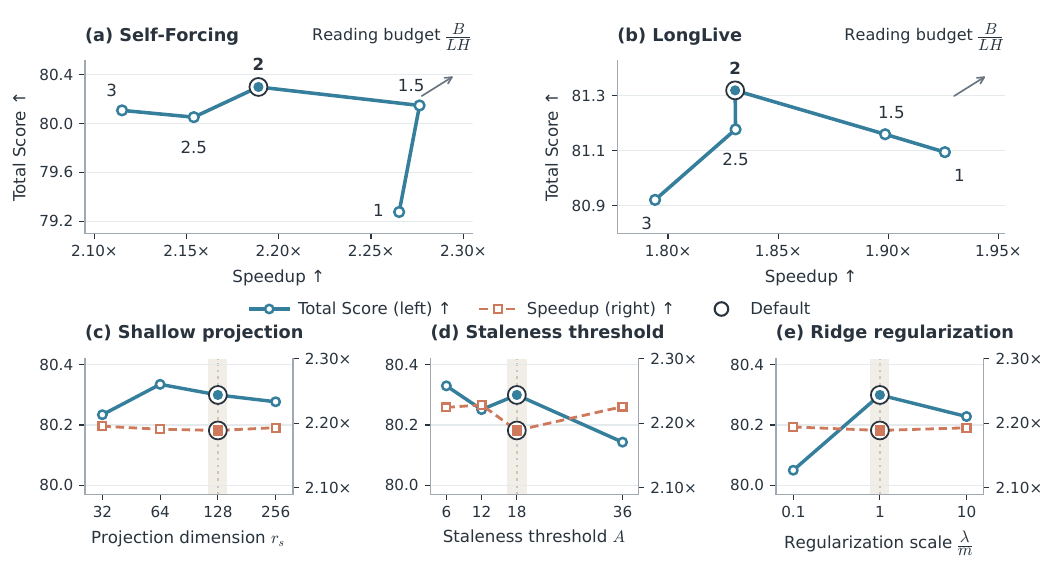}
\caption{Hyperparameter analysis of ScoutCache. (a,b) Tradeoff between Total Score and speedup under different normalized historical reading budgets $B/(LH)$ on Self-Forcing and LongLive. (c-e) Effects of shallow feature projection dimension $r_s$, staleness threshold $A$, and regularization scale $\lambda/m$ on generation quality and speedup using Self-Forcing.}
\label{fig:hyperparameters}
\end{figure*}

Figure~\ref{fig:hyperparameters} examines the effects of key hyperparameters on generation quality and inference efficiency. As shown in Figure~\ref{fig:hyperparameters} (a,b), increasing the historical reading budget generally improves quality at the cost of inference speed, while larger budgets do not necessarily yield further gains. The default budget $B=2LH$ achieves the highest Total Score on both backbones with a favorable speedup. Figure~\ref{fig:hyperparameters} (c-e) further evaluates the shallow feature projection dimension $r_s$, staleness threshold $A$, and regularization scale $\lambda/m$. Performance remains relatively stable across different projection dimensions and staleness thresholds, while $\lambda/m=1$ achieves the best quality. These results demonstrate the robustness of ScoutCache to moderate hyperparameter variations.

\section{Runtime Breakdown}

\begin{table}[t]
\centering
\caption{Breakdown of diffusion generation time across different video durations on Self-Forcing and LongLive. All times are measured in seconds. Denoising includes historical KV reading, while Other includes online calibration and remaining overhead.}
\label{tab:diffusion_breakdown}
\footnotesize
\setlength{\tabcolsep}{3pt}
\renewcommand{\arraystretch}{0.95}
\resizebox{\columnwidth}{!}{
\begin{tabular}{lcccc}
\toprule
Method & Denoising$\downarrow$ & KV Build$\downarrow$ & Other$\downarrow$ & Total$\downarrow$ \\
\midrule

\multicolumn{5}{l}{\textit{VBench (5s) -- Self-Forcing}} \\
Dummy Forcing & 5.132 & 1.297 & 0.050 & 6.48 \\
Forcing-KV & 5.155 & 1.327 & 0.027 & 6.51 \\
ScoutCache & \textbf{4.385} & \textbf{1.167} & 0.090 & \textbf{5.64} \\

\addlinespace[2pt]
\multicolumn{5}{l}{\textit{VBench (5s) -- LongLive}} \\
Dummy Forcing & 5.052 & 1.278 & 0.104 & 6.43 \\
Forcing-KV & 5.173 & 1.302 & 0.027 & 6.50 \\
ScoutCache & \textbf{4.384} & \textbf{1.170} & 0.087 & \textbf{5.64} \\

\midrule
\multicolumn{5}{l}{\textit{MovieGen (30s) -- Self-Forcing}} \\
Dummy Forcing & 30.685 & 7.840 & 0.056 & 38.58 \\
Forcing-KV & 31.997 & 8.403 & 0.032 & 40.43 \\
ScoutCache & \textbf{23.440} & \textbf{6.451} & 0.094 & \textbf{29.99} \\

\addlinespace[2pt]
\multicolumn{5}{l}{\textit{MovieGen (30s) -- LongLive}} \\
Dummy Forcing & 29.944 & 7.646 & 0.108 & 37.70 \\
Forcing-KV & 32.146 & 8.427 & 0.033 & 40.61 \\
ScoutCache & \textbf{23.426} & \textbf{6.443} & 0.090 & \textbf{29.96} \\

\midrule
\multicolumn{5}{l}{\textit{MovieGen (60s) -- Self-Forcing}} \\
Dummy Forcing & 61.076 & 15.589 & 0.061 & 76.73 \\
Forcing-KV & 63.382 & 16.614 & 0.038 & 80.03 \\
ScoutCache & \textbf{45.891} & \textbf{12.653} & 0.109 & \textbf{58.65} \\

\addlinespace[2pt]
\multicolumn{5}{l}{\textit{MovieGen (60s) -- LongLive}} \\
Dummy Forcing & 59.586 & 15.279 & 0.113 & 74.98 \\
Forcing-KV & 64.262 & 16.867 & 0.039 & 81.17 \\
ScoutCache & \textbf{45.774} & \textbf{12.635} & 0.095 & \textbf{58.50} \\

\midrule
\multicolumn{5}{l}{\textit{MovieGen (120s) -- Self-Forcing}} \\
Dummy Forcing & 121.452 & 31.058 & 0.073 & 152.58 \\
Forcing-KV & 126.481 & 33.201 & 0.051 & 159.73 \\
ScoutCache & \textbf{90.581} & \textbf{25.077} & 0.124 & \textbf{115.78} \\

\addlinespace[2pt]
\multicolumn{5}{l}{\textit{MovieGen (120s) -- LongLive}} \\
Dummy Forcing & 118.522 & 30.393 & 0.126 & 149.04 \\
Forcing-KV & 127.876 & 33.687 & 0.051 & 161.61 \\
ScoutCache & \textbf{90.598} & \textbf{25.075} & 0.105 & \textbf{115.78} \\

\bottomrule
\end{tabular}
}
\end{table}

\begin{table}[t]
\centering
\caption{Comparison of KV cache construction strategies. Latency is measured in milliseconds.}
\label{tab:cache_construction}
\small
\setlength{\tabcolsep}{5pt}
\begin{tabular}{lccc}
\toprule
Strategy & Tokens$\downarrow$ & Self-Forcing$\downarrow$ & LongLive$\downarrow$ \\
\midrule
Dense Construction & 4,680 & 194.55 & 194.06 \\
Post Compression & 4,680 & 201.41 & 201.25 \\
CCSC (Ours) & \textbf{3,510} & \textbf{151.23} & \textbf{151.27} \\
\bottomrule
\end{tabular}
\end{table}


\noindent\textbf{Diffusion Time Breakdown.}
To identify the sources of acceleration, we decompose the complete diffusion generation time into denoising, KV cache construction, and other overhead. As shown in Table~\ref{tab:diffusion_breakdown}, ScoutCache consistently reduces both denoising and cache construction costs across different video durations and backbones. Compared with Dummy Forcing, denoising time decreases by 25.4\% and 23.6\%, while KV cache construction time decreases by 19.3\% and 17.5\% for 120s generation on Self-Forcing and LongLive, respectively. Consequently, the total latency drops to 115.78s on both backbones, with minimal calibration and other overhead. Similar improvements are observed at shorter durations, demonstrating that ScoutCache achieves consistent efficiency gains by reducing computation in both major stages of diffusion generation.

\noindent\textbf{KV Cache Construction.}
We further isolate the cost of KV cache construction to evaluate the effectiveness of CCSC. As shown in Table~\ref{tab:cache_construction}, CCSC reduces the number of processed tokens from 4,680 to 3,510, decreasing construction latency by approximately 22\% on both backbones. In contrast, compressing the KV cache after dense construction still requires processing all tokens and introduces additional overhead. Compared with this approach, CCSC reduces construction latency by approximately 25\%, demonstrating the advantage of selecting spatial positions before Transformer processing.


\begin{figure}[t]
\centering
\includegraphics[width=0.85\columnwidth]{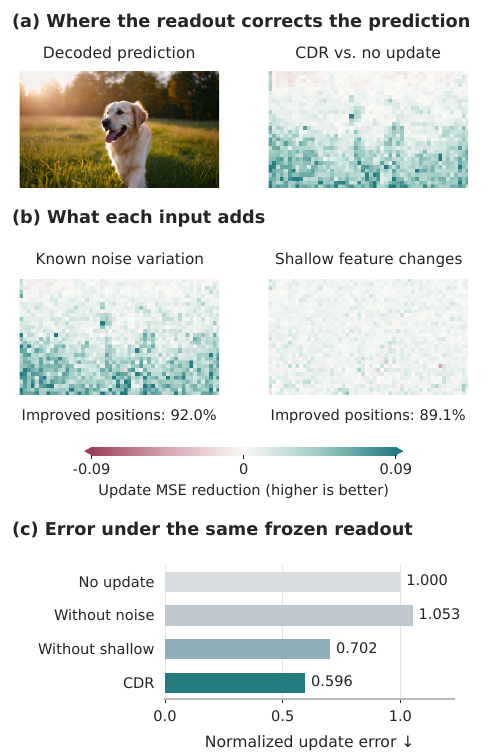}
\caption{Mechanism analysis of CDR. (a) Spatial distribution of prediction error reduction compared with no update. (b) Contributions of known noise variations and shallow feature changes to error reduction. (c) Normalized update errors under different input configurations with the same frozen readout, showing the benefit of combining both inputs.}
\label{fig:ma1}
\end{figure}

\begin{figure}[t]
\centering
\includegraphics[width=0.85\columnwidth]{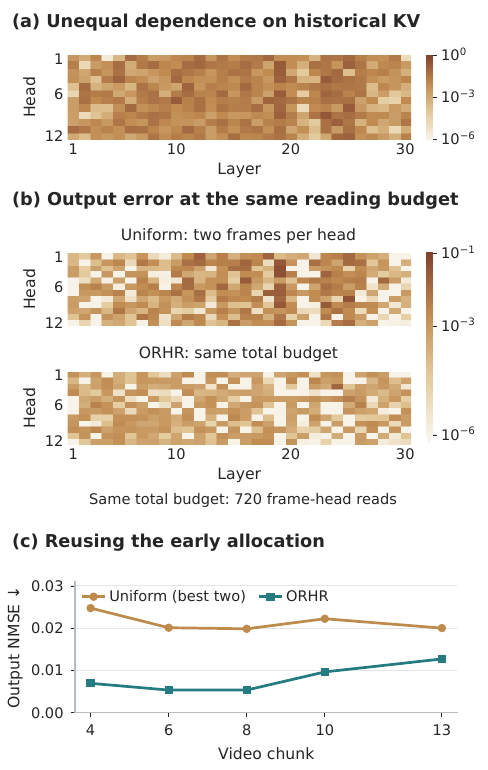}
\caption{Mechanism analysis of ORHR. (a) Unequal historical KV dependence across Transformer layers and heads. (b) Attention output errors under uniform reading and ORHR with the same budget of 720 frame accesses. (c) Output NMSE across subsequent video chunks using the early calibrated allocation, demonstrating consistently lower errors without recalibration.}
\label{fig:ma2}
\end{figure}

\begin{figure}[t]
\centering
\includegraphics[width=0.85\columnwidth]{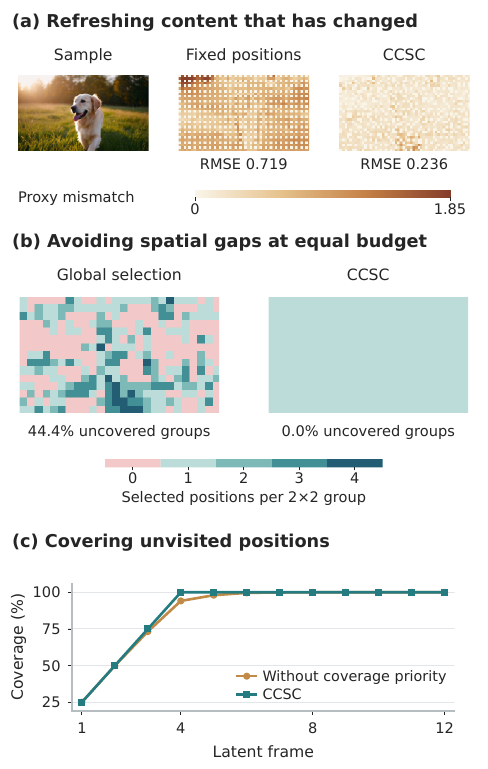}
\caption{Mechanism analysis of CCSC. (a) Proxy mismatch under fixed spatial positions and content guided selection. (b) Spatial coverage under equal selection budgets, showing how spatial grouping avoids uncovered regions. (c) Accumulated coverage across latent frames, demonstrating how coverage priority refreshes previously unvisited positions.}
\label{fig:ma3}
\end{figure}

\begin{figure*}[t]
\centering
\includegraphics[width=0.7\textwidth]{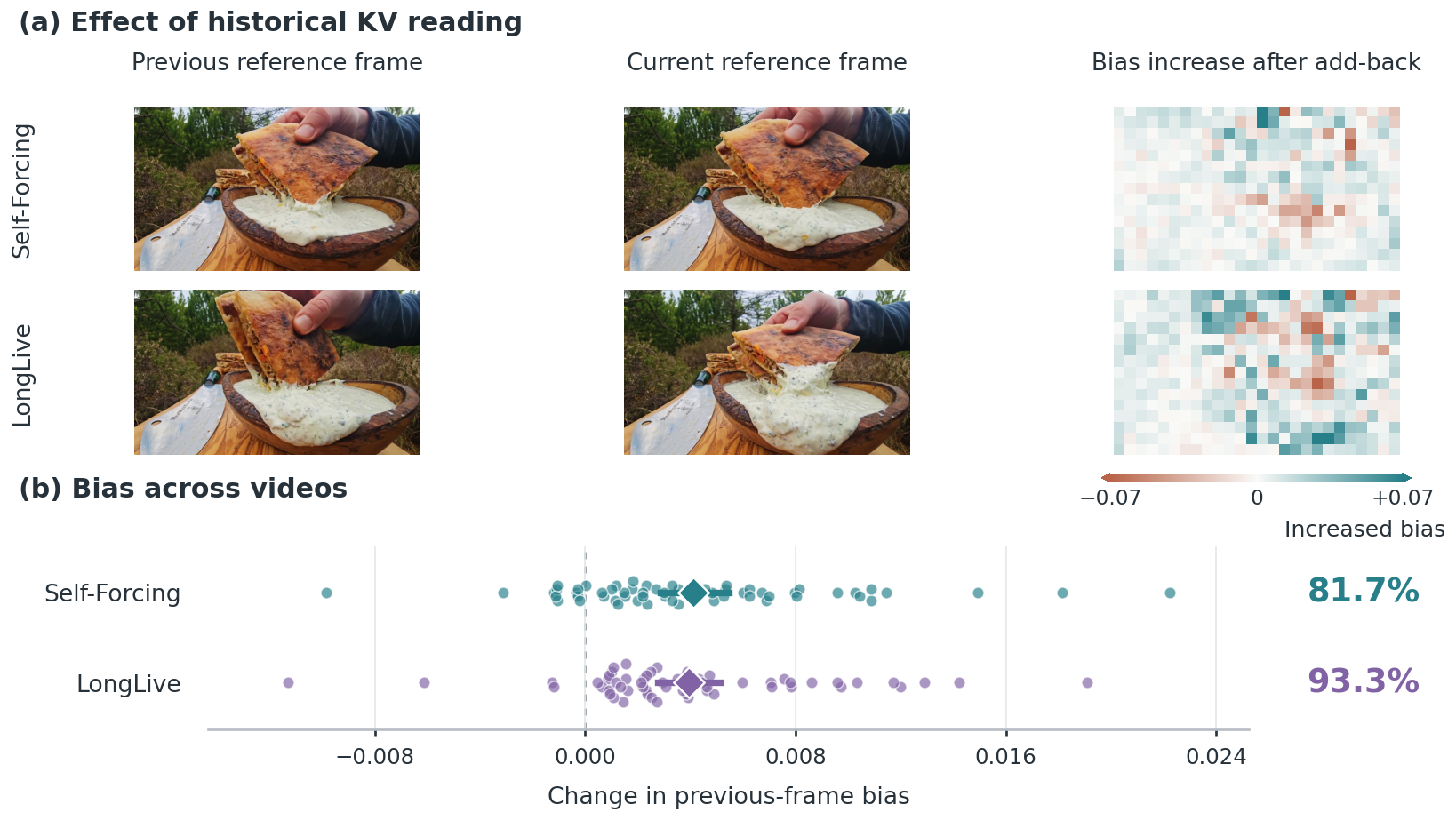}
\caption{Analysis of historical KV reading effects on generation quality. (a) Consecutive reference frames and spatial changes in previous-frame bias after restoring omitted historical KV reads. Blue and orange indicate increased and decreased bias, respectively. (b) Distribution of bias changes on Self-Forcing and LongLive, with percentages indicating the proportion of videos showing increased bias.}
\label{fig:ma4}
\end{figure*}

\section{Mechanism Analysis}

\noindent\textbf{Conditional Denoising Readout.} Figure~\ref{fig:ma1} examines how CDR estimates clean latent updates from known noise variations and shallow feature changes. The spatial visualizations show that both inputs contribute to reducing prediction errors at different positions. Compared with using no update, CDR reduces the normalized update error from 1.000 to 0.596. Removing either input increases the error, with the noise variation term showing a particularly important contribution. These results demonstrate that combining both inputs enables more accurate prediction updates using a fixed readout mapping.

\noindent\textbf{Historical KV Reading.} 
Figure~\ref{fig:ma2} analyzes the historical attention behavior and reading policy of ORHR. Historical KV dependence varies considerably across Transformer layers and heads, motivating nonuniform budget allocation. Under the same global budget of 720 frame accesses, ORHR produces lower attention output errors than uniform reading. Furthermore, the allocation calibrated on an early video chunk maintains lower output NMSE across subsequent chunks without recalibration. This supports the effectiveness of output risk guided allocation and its reuse during long video generation.

\noindent\textbf{Sparse Cache Construction.} Figure~\ref{fig:ma3} illustrates how content changes and spatial coverage guide CCSC. Compared with fixed spatial selection, CCSC reduces proxy mismatch RMSE from 0.719 to 0.236, indicating more effective updates of changing content. Under equal selection budgets, global selection leaves 44.4\% of spatial groups uncovered, while CCSC maintains complete group coverage. The coverage priority also accelerates the refresh of previously unvisited positions across latent frames. These results demonstrate that CCSC balances content relevance and spatial coverage during sparse KV cache construction.

\noindent\textbf{Generation Quality Improvement Analysis.}
Training-free acceleration methods often sacrifice generation quality for faster inference, while ScoutCache achieves improvements in both aspects. To understand this behavior, we further investigate the effect of historical KV reading on generation quality. The ablation results indicate that ORHR plays a major role in the quality improvements. 
As shown in Fig.~\ref{fig:ma4}(a), the reference frames reveal changes in object positions and poses between consecutive video chunks. However, the blue regions indicate that restoring full historical KV reading shifts current predictions toward the previous frame. This tendency may preserve outdated object positions and hinder motion progression. Fig.~\ref{fig:ma4}(b) further shows that previous-frame bias increases in 81.7\% and 93.3\% of videos on Self-Forcing and LongLive, respectively.
These findings suggest that selective historical KV reading can reduce interference from outdated content while preserving useful temporal context, helping explain the simultaneous efficiency and quality improvements of ScoutCache.

\section{Additional Qualitative Comparisons}

\begin{figure*}[t]
\centering
\includegraphics[width=0.9\textwidth]{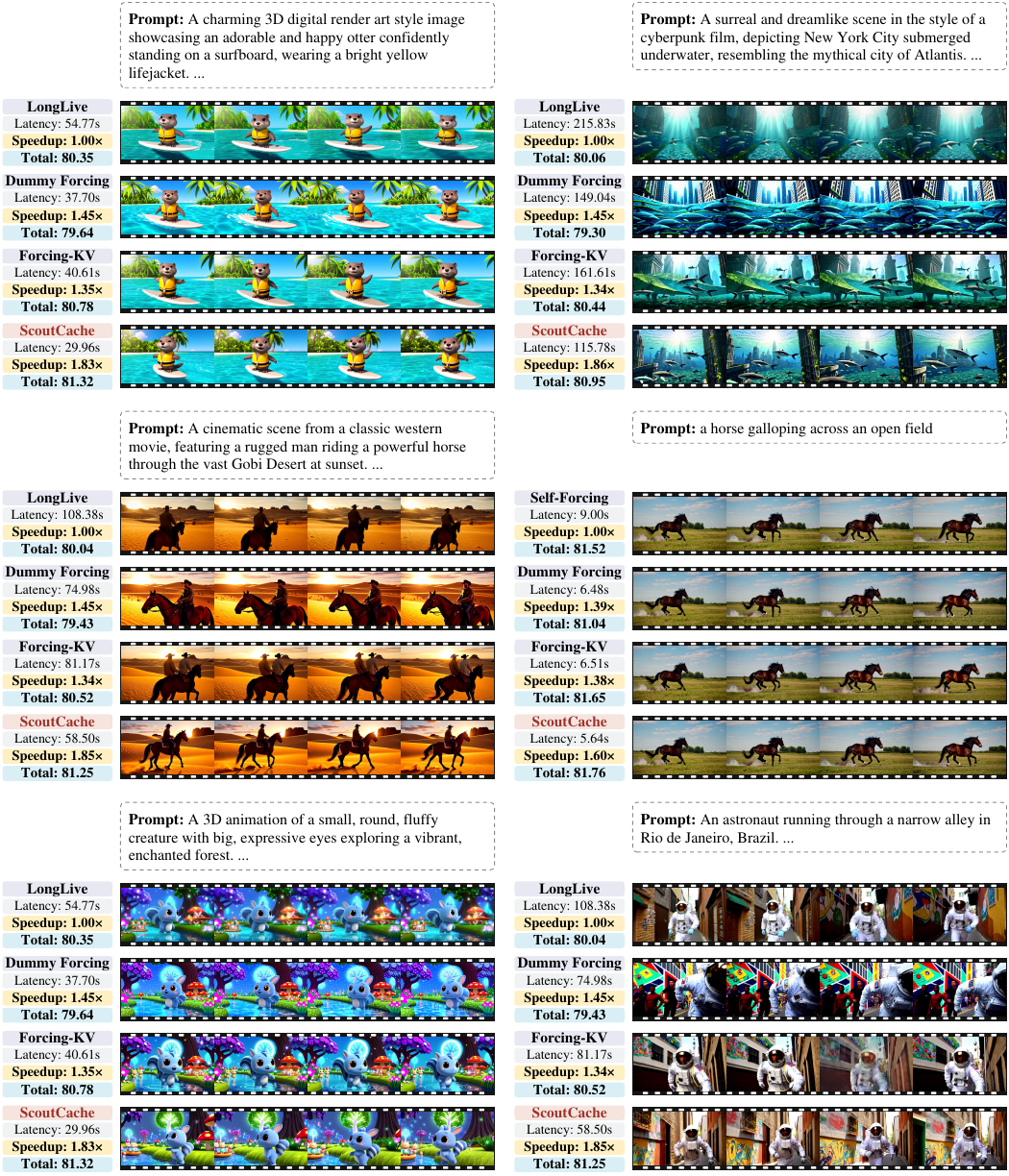}
\caption{Additional qualitative comparison of video generation results.}
\label{fig:additional_qualitative}
\end{figure*}

Figure~\ref{fig:additional_qualitative} presents additional qualitative comparisons across diverse scenes and video durations. In the underwater city scene, ScoutCache better separates the foreground fish from the surrounding architecture, preserving the spatial layout and depth of the scene. The surfing otter example further shows that ScoutCache maintains a consistent subject appearance while capturing changes in pose and movement.
Additional examples demonstrate the ability of ScoutCache to preserve visual details and motion in dynamic scenes. In the desert horseback riding scene, ScoutCache retains the rider and horse with visible leg movements, while maintaining consistent sunset lighting and background scenery. Similarly, the galloping horse sequence exhibits distinct running poses across frames, with stable subject appearance and surroundings. In the enchanted forest scene, ScoutCache preserves the character's appearance and rich environmental details while capturing changes in body orientation and movement. The astronaut running example provides a particularly clear comparison. Dummy Forcing exhibits noticeable changes in helmet appearance and scene composition, while Forcing-KV introduces ghosting artifacts in some frames. In contrast, ScoutCache maintains more consistent helmet and suit details, along with a clearer character silhouette, while preserving the running motion. Together, these examples further demonstrate that ScoutCache maintains visual fidelity and dynamic content while substantially reducing inference costs.

\section{Detailed Video Quality Results}
Tables~\ref{tab:sf_detailed_quality} and
\ref{tab:longlive_detailed_quality} report detailed video
quality scores on Self-Forcing and LongLive, respectively.
For 5s videos, we report the VBench Quality and Semantic
Scores, while longer videos are evaluated using six
VBench-Long dimensions. On Self-Forcing, ScoutCache achieves
the highest Total Score across all four video durations.
For 5s generation, it also obtains the highest Quality
Score of 84.38. On MovieGen, ScoutCache achieves the highest
Dynamic Degree at 30s and 60s, reaching 54.51 and 54.16,
respectively. It also consistently achieves the best
Imaging Quality across all three durations, demonstrating
its ability to retain visual details during accelerated
generation.
On LongLive, ScoutCache improves the 5s Total Score over
the original backbone and achieves the highest Total Score
across all MovieGen durations. Notably, it consistently
outperforms competing methods in Subject Consistency,
Dynamic Degree, and Imaging Quality for 30s, 60s, and 120s
generation. These improvements indicate that ScoutCache
preserves subject appearance while maintaining dynamic
content and visual fidelity. Although some baselines
achieve higher scores in individual dimensions, such as
Background Consistency and Motion Smoothness, ScoutCache
achieves the strongest overall video quality across both
backbones on longer videos. These results further support
the effectiveness of ScoutCache in accelerating
autoregressive video diffusion while preserving generation
quality.

\begin{table*}[t]
\centering
\caption{Detailed video quality scores on Self-Forcing.
The 5s results report the VBench Quality, Semantic, and
Total scores. The 30s, 60s, and 120s results report
six VBench-Long dimensions and the Total Score.}
\label{tab:sf_detailed_quality}
\footnotesize
\setlength{\tabcolsep}{3pt}
\renewcommand{\arraystretch}{1.06}
\begin{tabular}{@{}l*{10}{c}@{}}
\toprule
\raisebox{4pt}[0pt][0pt]{\textbf{Metric}}
& \shortstack{Self-\\Forcing}
& \shortstack{Flow\\Cache}
& \shortstack{Tea\\Cache}
& \shortstack{MotionCache\\(Slow)}
& \shortstack{MotionCache\\(Fast)}
& \raisebox{4pt}[0pt][0pt]{SVG2}
& \shortstack{Light\\Forcing}
& \shortstack{Dummy\\Forcing}
& \shortstack{Forcing-\\KV}
& \raisebox{4pt}[0pt][0pt]{\textbf{ScoutCache}} \\
\midrule

\multicolumn{11}{@{}l}{\textbf{VBench (5s)}} \\
Quality Score
& 84.06 & 84.04 & 83.61 & 72.96 & 72.73
& 83.41 & 83.44 & 83.56 & 84.34 & 84.38 \\
Semantic Score
& 71.36 & 71.64 & 70.92 & 52.75 & 43.40
& 70.70 & 71.66 & 70.95 & 70.92 & 71.28 \\
\cmidrule{1-11}
\textbf{Total Score}
& 81.52 & 81.56 & 81.07 & 68.92 & 66.86
& 80.87 & 81.08 & 81.04 & \underline{81.65} & \textbf{81.76} \\

\addlinespace[2pt]
\midrule
\multicolumn{11}{@{}l}{\textbf{MovieGen (30s)}} \\
Subject Consistency
& 96.74 & 96.15 & 96.36 & 92.43 & 91.18
& 95.16 & 96.47 & 96.42 & 95.91 & 96.01 \\
Background Consistency
& 96.14 & 95.86 & 95.96 & 94.47 & 93.73
& 94.88 & 96.10 & 96.08 & 95.29 & 95.51 \\
Motion Smoothness
& 98.58 & 98.35 & 98.68 & 95.43 & 94.79
& 98.52 & 98.68 & 98.21 & 98.14 & 98.08 \\
Dynamic Degree
& 39.01 & 40.63 & 36.97 & 18.02 & 46.04
& 44.69 & 34.01 & 48.33 & 51.25 & 54.51 \\
Aesthetic Quality
& 60.11 & 59.50 & 59.19 & 49.66 & 44.81
& 56.58 & 59.52 & 61.65 & 60.86 & 60.90 \\
Imaging Quality
& 68.53 & 66.57 & 67.69 & 51.76 & 45.91
& 63.27 & 67.83 & 69.56 & 68.85 & 69.98 \\
\cmidrule{1-11}
\textbf{Total Score}
& 79.10 & 78.45 & 78.54 & 68.97 & 68.72
& 77.35 & 78.47 & \underline{80.11} & 79.76 & \textbf{80.30} \\

\addlinespace[2pt]
\midrule
\multicolumn{11}{@{}l}{\textbf{MovieGen (60s)}} \\
Subject Consistency
& 97.31 & 96.18 & 96.93 & 93.61 & 91.78
& 94.91 & 96.83 & 96.50 & 95.49 & 97.24 \\
Background Consistency
& 96.67 & 96.08 & 96.46 & 95.03 & 93.96
& 95.10 & 96.15 & 96.02 & 94.94 & 95.41 \\
Motion Smoothness
& 98.57 & 98.05 & 98.60 & 95.46 & 94.93
& 98.03 & 98.59 & 98.03 & 97.93 & 97.96 \\
Dynamic Degree
& 29.79 & 33.77 & 26.53 & 13.62 & 47.29
& 39.50 & 31.66 & 45.59 & 51.22 & 54.16 \\
Aesthetic Quality
& 57.57 & 55.82 & 56.14 & 47.99 & 43.76
& 53.97 & 58.52 & 61.08 & 59.94 & 58.88 \\
Imaging Quality
& 66.88 & 64.16 & 67.52 & 52.01 & 44.03
& 61.48 & 67.69 & 68.78 & 68.73 & 68.91 \\
\cmidrule{1-11}
\textbf{Total Score}
& 77.76 & 76.60 & 77.20 & 68.71 & 68.57
& 75.77 & 78.03 & \underline{79.51} & 79.26 & \textbf{79.87} \\

\addlinespace[2pt]
\midrule
\multicolumn{11}{@{}l}{\textbf{MovieGen (120s)}} \\
Subject Consistency
& 96.54 & 96.35 & 96.11 & 94.00 & 92.16
& 95.13 & 97.15 & 96.22 & 95.67 & 97.29 \\
Background Consistency
& 96.66 & 96.34 & 96.34 & 95.28 & 94.11
& 95.62 & 96.42 & 95.90 & 94.97 & 95.43 \\
Motion Smoothness
& 97.99 & 98.14 & 97.97 & 95.36 & 95.20
& 97.56 & 98.59 & 97.86 & 98.00 & 98.09 \\
Dynamic Degree
& 28.24 & 28.68 & 25.36 & 17.57 & 55.65
& 37.94 & 30.22 & 44.24 & 45.80 & 49.14 \\
Aesthetic Quality
& 51.27 & 51.14 & 51.07 & 45.66 & 41.52
& 49.72 & 56.90 & 60.26 & 59.14 & 56.77 \\
Imaging Quality
& 61.85 & 59.64 & 63.05 & 51.04 & 42.40
& 58.81 & 67.00 & 68.68 & 68.63 & 68.82 \\
\cmidrule{1-11}
\textbf{Total Score}
& 75.03 & 74.62 & 74.77 & 68.56 & 68.91
& 74.25 & 77.62 & \underline{79.02} & 78.70 & \textbf{79.11} \\

\bottomrule
\end{tabular}
\end{table*}

\begin{table*}[t]
\centering
\caption{Detailed video quality scores on LongLive.
The 5s results report the VBench Quality, Semantic, and
Total scores. The 30s, 60s, and 120s results report
six VBench-Long dimensions and the Total Score.}
\label{tab:longlive_detailed_quality}
\footnotesize
\setlength{\tabcolsep}{3pt}
\renewcommand{\arraystretch}{1.06}
\begin{tabular}{@{}l*{7}{c}@{}}
\toprule
\raisebox{4pt}[0pt][0pt]{\textbf{Metric}}
& \raisebox{4pt}[0pt][0pt]{LongLive}
& \shortstack{Flow\\Cache}
& \shortstack{Tea\\Cache}
& \raisebox{4pt}[0pt][0pt]{SVG2}
& \shortstack{Dummy\\Forcing}
& \shortstack{Forcing-\\KV}
& \raisebox{4pt}[0pt][0pt]{\textbf{ScoutCache}} \\
\midrule

\multicolumn{8}{@{}l}{\textbf{VBench (5s)}} \\
Quality Score
& 81.83 & 81.78 & 82.36 & 82.03 & 81.55 & 82.66 & 82.14 \\
Semantic Score
& 70.84 & 71.36 & 71.23 & 70.78 & 71.25 & 71.28 & 70.71 \\
\cmidrule{1-8}
\textbf{Total Score}
& 79.63 & 79.70 & \underline{80.14} & 79.79
& 79.50 & \textbf{80.39} & 79.85 \\

\addlinespace[2pt]
\midrule
\multicolumn{8}{@{}l}{\textbf{MovieGen (30s)}} \\
Subject Consistency
& 97.59 & 97.46 & 97.54 & 96.91 & 97.92 & 97.58 & 98.04 \\
Background Consistency
& 96.73 & 96.57 & 96.62 & 96.02 & 97.17 & 96.67 & 96.59 \\
Motion Smoothness
& 98.82 & 98.73 & 98.82 & 98.78 & 98.88 & 98.64 & 98.60 \\
Dynamic Degree
& 42.81 & 41.35 & 38.80 & 45.72 & 26.92 & 45.15 & 48.15 \\
Aesthetic Quality
& 61.94 & 62.07 & 61.81 & 59.96 & 62.70 & 62.44 & 62.82 \\
Imaging Quality
& 69.00 & 68.86 & 68.81 & 66.00 & 71.08 & 70.39 & 71.21 \\
\cmidrule{1-8}
\textbf{Total Score}
& 80.35 & 80.09 & 79.89 & 79.37
& 79.64 & \underline{80.78} & \textbf{81.32} \\

\addlinespace[2pt]
\midrule
\multicolumn{8}{@{}l}{\textbf{MovieGen (60s)}} \\
Subject Consistency
& 97.38 & 97.23 & 97.23 & 96.54 & 97.94 & 97.48 & 98.06 \\
Background Consistency
& 96.61 & 96.49 & 96.45 & 95.79 & 97.23 & 96.54 & 96.39 \\
Motion Smoothness
& 98.76 & 98.72 & 98.70 & 98.67 & 98.81 & 98.54 & 98.44 \\
Dynamic Degree
& 39.84 & 41.84 & 37.42 & 47.37 & 24.16 & 44.27 & 50.22 \\
Aesthetic Quality
& 62.34 & 62.07 & 61.80 & 60.04 & 62.85 & 62.14 & 61.97 \\
Imaging Quality
& 68.97 & 68.46 & 68.81 & 66.26 & 71.31 & 70.34 & 71.45 \\
\cmidrule{1-8}
\textbf{Total Score}
& 80.04 & 79.99 & 79.58 & 79.37
& 79.43 & \underline{80.52} & \textbf{81.25} \\

\addlinespace[2pt]
\midrule
\multicolumn{8}{@{}l}{\textbf{MovieGen (120s)}} \\
Subject Consistency
& 97.31 & 97.18 & 97.14 & 96.52 & 97.89 & 97.48 & 98.00 \\
Background Consistency
& 96.52 & 96.39 & 96.34 & 95.68 & 97.24 & 96.53 & 96.24 \\
Motion Smoothness
& 98.74 & 98.70 & 98.70 & 98.66 & 98.80 & 98.53 & 98.37 \\
Dynamic Degree
& 42.24 & 41.38 & 39.57 & 48.32 & 23.25 & 44.23 & 49.60 \\
Aesthetic Quality
& 61.59 & 61.80 & 61.45 & 59.95 & 62.30 & 61.74 & 60.84 \\
Imaging Quality
& 68.93 & 68.60 & 68.91 & 66.33 & 71.67 & 70.40 & 71.78 \\
\cmidrule{1-8}
\textbf{Total Score}
& 80.06 & 79.88 & 79.68 & 79.42
& 79.30 & \underline{80.44} & \textbf{80.95} \\
\bottomrule
\end{tabular}
\end{table*}

\end{document}